# Levels of Autonomous Radiology

**Suraj Ghuwalewala[1], Viraj Kulkarni[1], Richa Pant[1,*], and Amit Kharat[1,2]**

[1] DeepTek, Pune, India
[2] Dr. D.Y. Patil Medical College, Hospital and Research Centre, Pune, India
[*] richa.pant@deeptek.ai

## Abstract

Radiology, being one of the younger disciplines of medicine with a history of just over a century, has witnessed tremendous technological advancements and has revolutionized the way we practice medicine today. In the last few decades, medical imaging modalities have generated seismic amounts of medical data. The development and adoption of Artificial Intelligence (AI) applications using this data will lead to the next phase of evolution in radiology. It will include automating laborious manual tasks such as annotations, report-generation, etc., along with the initial radiological assessment of cases to aid radiologists in their evaluation workflow. We propose a level-wise classification for the progression of automation in radiology, explaining AI assistance at each level with corresponding challenges and solutions. We hope that such discussions can help us address the challenges in a structured way and take the necessary steps to ensure the smooth adoption of new technologies in radiology.

## Introduction

The Fifth Industrial Revolution (5IR) is upon us, as evidenced by the ever-increasing use and worldwide adoption of AI applications, especially those pertaining to image-based and text-based tasks [1]. This is due to the availability of massive amounts of data, often referred to as "Big Data", extraordinary advances in computing power, and the continuous development of new deep learning algorithms over the last decade.

In radiology, these AI applications are being widely adopted for assisted image acquisition, post-processing, automated diagnosis, and report generation. Automation in this field is still in its infancy, and several clinical and ethical challenges must be addressed before further progress can be made [2].

The automotive industry is another sector that is witnessing such extensive adoption of AI [3]. The Society of Automotive Engineers (SAE) has classified the progression of driving automation into six levels [4], ranging from No Automation (Level 0) to Full Automation (Level 5). As vehicle autonomy increases with each level of automation, driver intervention is reduced.

In this comment, we attempt to categorize and map the advancements and challenges of automation in radiology into six levels, similar to driving automation. The subsequent part of the article briefly discusses each level, its challenges, plausible solutions, and enabling factors for transitioning into the next level.

## Levels of automation in radiology

The advancement of artificial intelligence in the health sector has significantly bridged the gap between computation and radiology, paving the way for automation in radiology practice. We describe the six levels of automation in radiology using a taxonomy similar to that used in driving automation. We further attempt to provide a futuristic vision of the challenges that the radiology field may encounter as we progress towards the complete automation of this field. Figure 1 illustrates different levels of automation in radiology, including the challenges at each stage and the factors that enable the progression between levels.

### Level 0: No Automation

*Level 0,* also known as *No Automation*, is the stage where a radiologist manually performs every task from image acquisition and radiographic film processing to diagnostic analysis without the assistance of AI. We are well past this stage as the recent advances in medical imaging modalities have enabled digital storage and processing of the scans along with some automated assistance to aid in the imaging workflow.

### Level 1: Radiologist Assistance

At *Level 1* automation, a radiologist performs most tasks manually with assistance from machines. Recent technological advancements have digitized medical scans, making it easier for radiologists to store, maintain, and distribute the data. Furthermore, newer solutions include features such as contrast-brightness adjustment, assisted stitching of scans, assisted focus adjustment, etc. which simplify the imaging workflow and enable detailed radiological analysis. With everything digitized, these modalities generate enormous amounts of data, and the biggest challenge at this stage is proper maintenance and storage of data [5]. This is where technologies like Picture Archiving and Communication Systems (PACS) have provided an economical solution to compress and store data for easy retrieval and sharing [6]. We believe that the majority of radiological evaluations performed in clinical settings around the globe are at this stage of automation.

### Level 2: Partial Automation

Partial automation in radiology refers to the use of computer-assisted diagnostic modalities to automate prioritization. However, the automation at level 2 requires radiologist supervision, and the diagnostic decision is not final without the radiologist's approval. With the advancement of PACS technology, radiology practices frequently consider upgrades and renovations to further improve efficiency and prowess. For example, radiomics is an emerging subfield of machine learning that converts radiographic images into mineable high-

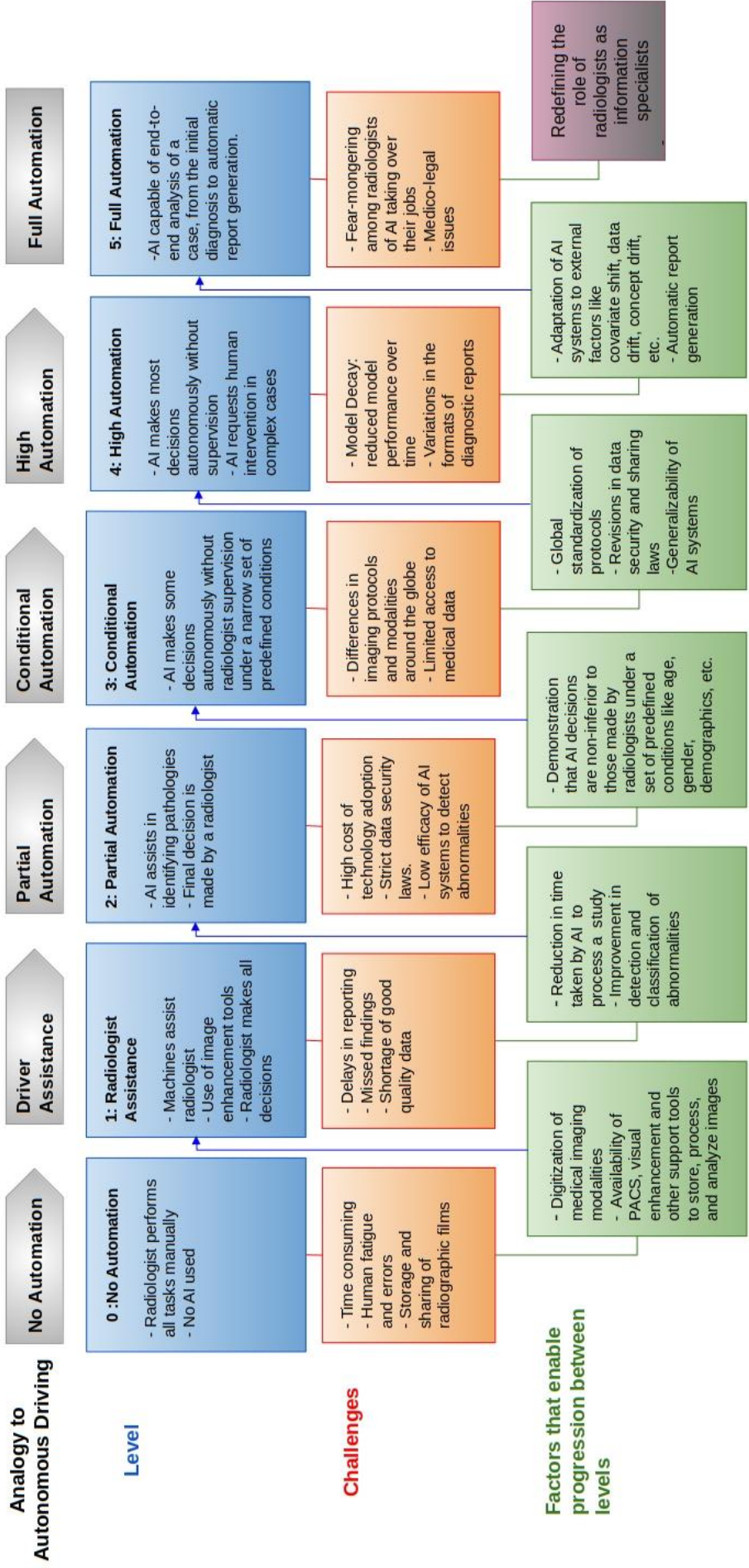

**Figure 1:** Flowchart depicting the various levels of automation in radiology practice. At each level, the role of the radiologist and AI is outlined, along with the enabling factors required to mitigate the potential challenges for progression to the next level.

-dimensional data. Machine learning algorithms can be employed to extract features from radiographic images that can help make prognostic decisions [7]. The amalgamation of machine learning and radiomics has the potential to improve its application in precision medicines and clinical practices. Since these technologies are still in their nascent stages of development, they should only be used for basic assistance and data processing, and the final decision must be made by a radiologist.

The progress at this level of automation is slow and can be attributed to three major factors: (a) the annotation and documented diagnosis of the medical data by an expert are time-consuming and expensive, resulting in a limited amount of good quality data for the development of AI systems [8], (b) medical data is often governed by several data security laws, regulations, and compliances, making it extremely difficult to share and use this data outside a clinical setting [2], (c) current algorithms for analyzing radiological scans are compute-hungry, significantly increasing the cost of adopting these technologies.

Collaborations between hospitals and tech companies are critical to bypass the barriers of data sharing laws and make the best use of rich medical data to develop advanced solutions for automated and accurate diagnosis. It is also important to develop low-power and cost-effective solutions that can be easily adopted by medical organizations.

**Level 3: Conditional Automation**

Unlike partial automation, where the final decision is entirely dependent on the radiologist, the systems at *Level 3: Conditional Automation* are robust enough to diagnose and make decisions under a predefined set of conditions without radiologist supervision. If these conditions are not met, a radiologist must be available to override the AI analysis. As previously stated, sharing of medical data outside of a clinical setting is governed by several laws and regulations, which makes cross-border sharing of this data even more difficult [2]. Additionally, imaging protocols, manufacturers of the imaging modalities, and the process of storing and processing medical data differ between organizations, which impede the use of data from different sources for AI applications [9]. These factors result in the development of AI systems on a limited distribution of data, making them highly susceptible to failure if certain conditions, such as demographics, race, gender, time, etc., are not met. For example, Dou *et al.* developed a COVID-19 detection model using datasets from internal hospitals in Hong Kong. The model performed extremely well in identifying abnormalities in Chinese datasets but underperformed in German datasets with different population demographics [10]. Cross-center training of the AI model for different demographics and distinct cohort features would help the model learn from multiple sources and mitigate the problem of generalizability.

**Level 4: High Automation**

Advancing from level 3, the AI systems at *Level 4: High Automation,* would be thoroughly tested and approved to analyze radiological scans and make decisions without the assistance of a radiologist. Human intervention would only be required in complex cases where AI requests it.

Most countries do not have an adequate number of radiology experts, which increases the workload on existing ones [11], causing delays in patient triage. The automation at this level will most likely be used to automatically triage and prioritize high-risk patients, and assist radiologists to analyze cases quicker, thus reducing the radiation dose and delays in diagnoses. It will also improve access to medical imaging in low-resource countries, allowing diagnostics to reach a larger population. Such systems should be reliable and accurate to assess cases under varying conditions such as different age groups, demographics, data acquisition protocols, etc., which will require medical data from multiple sources. Therefore, standardization of annotation, data processing, and storage protocols will be vital for this data to be efficiently used by tech companies around the world [12]. Imaging biobanks would be required for organizing and sharing multi-institutional image data from which AI models could be trained [13]. Additionally, the laws and regulations governing medical data security and sharing across countries and continents would need revisions with equal cooperation from tech companies to comply with the data anonymization and storage protocols. Governing bodies like FDA, EMA, etc. will have to play a huge role in making this a reality [9]. Techniques like federated learning allow AI models to be developed on multiple devices dispersed across geographical borders. This could be extremely useful when the data governing laws prohibit cross-border movement of data [14,15]. Lastly, adhering to these standardized protocols would incur additional costs that could be shared by both medical organizations and tech companies.

**Level 5: Full Automation**

*Level 5,* referred to as *Full Automation*, is the ultimate stage of automation in radiology, where an AI application would be capable of end-to-end analysis of a case, from the initial diagnosis to automatic report generation. With the recent advances in Natural Language Processing (NLP), technologies like Generative Pre-trained Transformer-3 (GPT-3) can be used for automatic and standardized diagnostic report generation [16]. The performance of AI systems tends to decay over time because of covariate shift [17], data drift [18], and concept drift [19], but these advanced systems would be resilient to such factors. They would adhere to the concept of continual learning where they constantly adapt to the ever-changing environment whilst not forgetting the previous learnings.

The complete automation of radiology in clinical practice will be challenged by medico-legal concerns about assigning liability in cases of AI misdiagnosis [13]. Another challenge at this stage would be to address the fear among radiologists of AI systems taking over their jobs [20]. However, jobs will not be lost, but rather roles will be redefined. With the influx of new data, radiologists would be the information specialists capable of piloting artificial intelligence and guiding medical data to improve patient care [21]. We believe that AI systems will become smart assistants for radiologists, capable of automatically performing mundane tasks such as preliminary diagnosis, annotations, report generation, etc. under radiologist supervision. This will not only reduce the workload of radiologists but will also allow them to actively participate in other aspects of patient care. The adoption of AI systems will significantly increase the number of early detections and reduce the number of misdiagnoses, leading to better overall patient care.

## Conclusion

The advancement in artificial intelligence is bringing the field of radiology to a higher level of automation. We propose a level-wise classification system for automation in radiology to elucidate the step-by-step adoption of AI in clinical practice. We also highlight the concerns and challenges that must be addressed as radiology advances toward complete automation. The development, deployment, and adoption of machine learning models in clinical practice necessitates addressing several important issues. Kulkarni *et al*. proposed a comprehensive list of considerations, such as the high cost of annotations, model generalization to unfamiliar data sets, model decay, fairness and bias, clinical validation, etc. that must be addressed for successful adoption of machine learning models in routine clinical practice [22]. With the ever-increasing amounts of medical data and advances in computation technologies, automation in radiology is imminent. The key factor is the time it takes to reach the stage of full automation, which depends on the collective efforts of medical experts and AI scientists, along with the acceptance and adoption of these technologies in clinical workflows.

The *"expert-in-loop"* ideology adopted by companies like DeepTek [23] aids in achieving synergy among AI scientists, software developers, and expert radiologists. This significantly improves the quality and quantity of expert clinical feedback and guidance at every stage of development. As we move closer to complete automation of radiological analysis, such collaborations are crucial for expediting the automation process.

# References

1    Sarfraz Z, Sarfraz A, Iftikar HM, Akhund R. Is COVID-19 pushing us to the Fifth Industrial Revolution (Society 5.0)? *Pak J Med Sci Q* 2021; **37**. DOI:10.12669/pjms.37.2.3387.

2    Recht MP, Dewey M, Dreyer K, *et al.* Integrating artificial intelligence into the clinical practice of radiology: challenges and recommendations. *Eur Radiol* 2020; **30**: 3576–84.

3    Khan AM, Bacchus A, Erwin S. Policy challenges of increasing automation in driving. IATSS Research. 2012; **35**: 79–89.

4    SAE MOBILUS. https://doi.org/10.4271/J3016_202104 (accessed Oct 20, 2021).

5    Aiello M, Cavaliere C, D'Albore A, Salvatore M. The challenges of diagnostic imaging in the era of Big data. *J Clin Med Res* 2019; **8**: 316.

6    Choplin RH, Boehme JM 2nd, Maynard CD. Picture archiving and communication systems: an overview. *Radiographics* 1992; **12**: 127–9.

7    Zhang B, He X, Ouyang F, *et al.* Radiomic machine-learning classifiers for prognostic biomarkers of advanced nasopharyngeal carcinoma. *Cancer Lett* 2017; **403**: 21–7.

8    Willemink MJ, Koszek WA, Hardell C, *et al.* Preparing Medical Imaging Data for Machine Learning. *Radiology* 2020; published online Feb 18. DOI:10.1148/radiol.2020192224.

9    Thrall JH, Li X, Li Q, *et al.* Artificial Intelligence and Machine Learning in Radiology: Opportunities, Challenges, Pitfalls, and Criteria for Success. *J Am Coll Radiol* 2018; **15**: 504–8.

10   Dou Q, So TY, Jiang M, *et al.* Federated deep learning for detecting COVID-19 lung abnormalities in CT: a privacy-preserving multinational validation study. *npj Digital Medicine* 2021; **4**: 1–11.

11   Bruls RJM, Kwee RM. Workload for radiologists during on-call hours: dramatic increase in the past 15 years. *Insights Imaging* 2020; **11**: 121.

12   Dewey M. The future of radiology: adding value to clinical care. Lancet. 2018; **392**: 472–3.

13   (esr) ES of R, European Society of Radiology (ESR). What the radiologist should know about artificial intelligence – an ESR white paper. Insights into Imaging. 2019; **10**. DOI:10.1186/s13244-019-0738-2.

14   Gawali M, Arvind CS, Suryavanshi S, *et al.* Comparison of privacy-preserving distributed deep learning methods in healthcare. In: Medical Image Understanding and Analysis. Cham: Springer International Publishing, 2021: 457–71.

15   Kulkarni V, Kulkarni M, Pant A. Survey of personalization techniques for federated

learning. In: 2020 Fourth World Conference on Smart Trends in Systems, Security and Sustainability (WorldS4). IEEE, 2020. DOI:10.1109/worlds450073.2020.9210355.

16  Korngiebel DM, Mooney SD. Considering the possibilities and pitfalls of Generative Pre-trained Transformer 3 (GPT-3) in healthcare delivery. *NPJ Digit Med* 2021; **4**: 93.

17  Covariate Shift: A Review and Analysis on Classifiers. https://doi.org/10.1109/GCAT47503.2019.8978471 (accessed Oct 21, 2021).

18  Ackerman S, Farchi E, Raz O, Zalmanovici M, Dube P. Detection of data drift and outliers affecting machine learning model performance over time. 2020; published online Dec 16. http://arxiv.org/abs/2012.09258 (accessed Oct 21, 2021).

19  Žliobaitė I. Learning under Concept Drift: an Overview. 2010; published online Oct 22. http://arxiv.org/abs/1010.4784 (accessed Oct 20, 2021).

20  Pakdemirli E. Artificial intelligence in radiology: friend or foe? Where are we now and where are we heading? *Acta Radiol Open* 2019; **8**: 2058460119830222.

21  Jha S, Topol EJ. Adapting to Artificial Intelligence: Radiologists and Pathologists as Information Specialists. *JAMA* 2016; **316**: 2353–4.

22  Kulkarni V, Gawali M, Kharat A. Key Technology Considerations in Developing and Deploying Machine Learning Models in Clinical Radiology Practice. *JMIR Med Inform* 2021; **9**: e28776.

23  Deeptek.ai. https://www.deeptek.ai (accessed Oct 21, 2021).